\documentclass[10pt,twocolumn,letterpaper]{article}

\usepackage{cvpr}              %
\usepackage{booktabs}

\usepackage[accsupp]{axessibility}
\usepackage{graphicx}
\usepackage[normalem]{ulem}
\usepackage{bbm}
\usepackage{multirow}

\let\svthefootnote\thefootnote
\newcommand\freefootnote[1]{%
  \let\thefootnote\relax%
  \footnotetext{#1}%
  \let\thefootnote\svthefootnote%
}
\definecolor{cvprblue}{rgb}{0.21,0.49,0.74}
\definecolor{mygray}{gray}{.9}
\usepackage[pagebackref,breaklinks,colorlinks,allcolors=cvprblue]{hyperref}
\usepackage{colortbl}

\newcommand{\DATASET}{VL-RewardBench}%

\definecolor{mydarkgreen}{RGB}{0, 100, 0}

\def\paperID{16268} %
\def\confName{CVPR}
\def\confYear{2025}

\title{\DATASET{}: A Challenging Benchmark for Vision-Language Generative Reward Models  
}
\author{Lei Li$^{1*}$ \quad Yuancheng Wei$^{2*}$ \quad Zhihui Xie$^{1*}$ \quad Xuqing Yang$^{3*}$ \quad Yifan Song$^{4}$  \quad Peiyi Wang$^{4}$  \\  Chenxin An$^{1}$  \quad
Tianyu Liu$^{4}$ \quad
Sujian Li$^{4}$ \quad Bill Yuchen Lin$^{5,6}$ \quad Lingpeng Kong$^{1}$ \quad Qi Liu$^{1}$\\ \\ 
$^{1}$HKU\quad 
$^{2}$SCUT\quad
$^{3}$SJTU\quad 
$^{4}$PKU\quad 
$^{5}$UW \quad $^{6}$Allen AI\\
{\tt\small \{nlp.lilei, wyc528813339, zhxieml, catalpabungeiyang\}@gmail.com}
}

\begin{document}
\maketitle
\begin{abstract}

Vision-language generative reward models (VL-GenRMs) play a crucial role in aligning and evaluating multimodal AI systems, yet their own evaluation remains under-explored. 
Current assessment methods primarily rely on AI-annotated preference labels from traditional VL tasks, which can introduce biases and often fail to effectively challenge state-of-the-art models.
To address these limitations, we introduce \DATASET, a comprehensive benchmark spanning general multimodal queries, visual hallucination detection, and complex reasoning tasks.
Through our AI-assisted annotation pipeline that combines sample selection with human verification, we curate 1,250 high-quality examples specifically designed to probe VL-GenRMs limitations.
Comprehensive evaluation across 16 leading large vision-language models demonstrates \DATASET{}'s effectiveness as a challenging testbed, where even GPT-4o achieves only 65.4\% accuracy, and state-of-the-art open-source models such as Qwen2-VL-72B, struggle to surpass random-guessing. 
Importantly, performance on \DATASET\ strongly correlates (Pearson's r $>$ 0.9) with MMMU-Pro accuracy using Best-of-N sampling with VL-GenRMs.
Analysis experiments uncover three critical insights for improving VL-GenRMs: (i) models predominantly fail at basic visual perception tasks rather than reasoning tasks; 
(ii) inference-time scaling benefits vary dramatically by model capacity; and (iii) training VL-GenRMs to learn to judge substantially boosts judgment capability (+14.7\% accuracy for a 7B VL-GenRM).
We believe \DATASET\ along with the experimental insights will become a valuable resource for advancing VL-GenRMs. Project page: \url{https://vl-rewardbench.github.io}.
\freefootnote{\hspace{-4pt}$^*$Core contributors.}

\end{abstract}

\section{Introduction}
\label{sec:intro}
Large vision-language models~(LVLMs) such as GPT-4o~\cite{gpt4o} have demonstrated remarkable capabilities across diverse multimodal perception and cognition tasks~\citep{fu2023mme,yue2023mmmu,videomme}. 
Building on these capabilities, LVLMs are increasingly being deployed as vision-language generative reward models (VL-GenRMs) to automatically assess model responses~\cite{gpt4v-evaluator,li2024vlfeedback,llava-critic}. This LVLM-as-a-judge paradigm has emerged as a scalable solution for model alignment, enabling efficient model ranking and high-quality sample selection~\citep{jing2024fgaif,yu2024rlaif,Zhou2024CalibratedSV}.

\begin{figure}[t]
    \centering
    \includegraphics[width=0.95\linewidth]{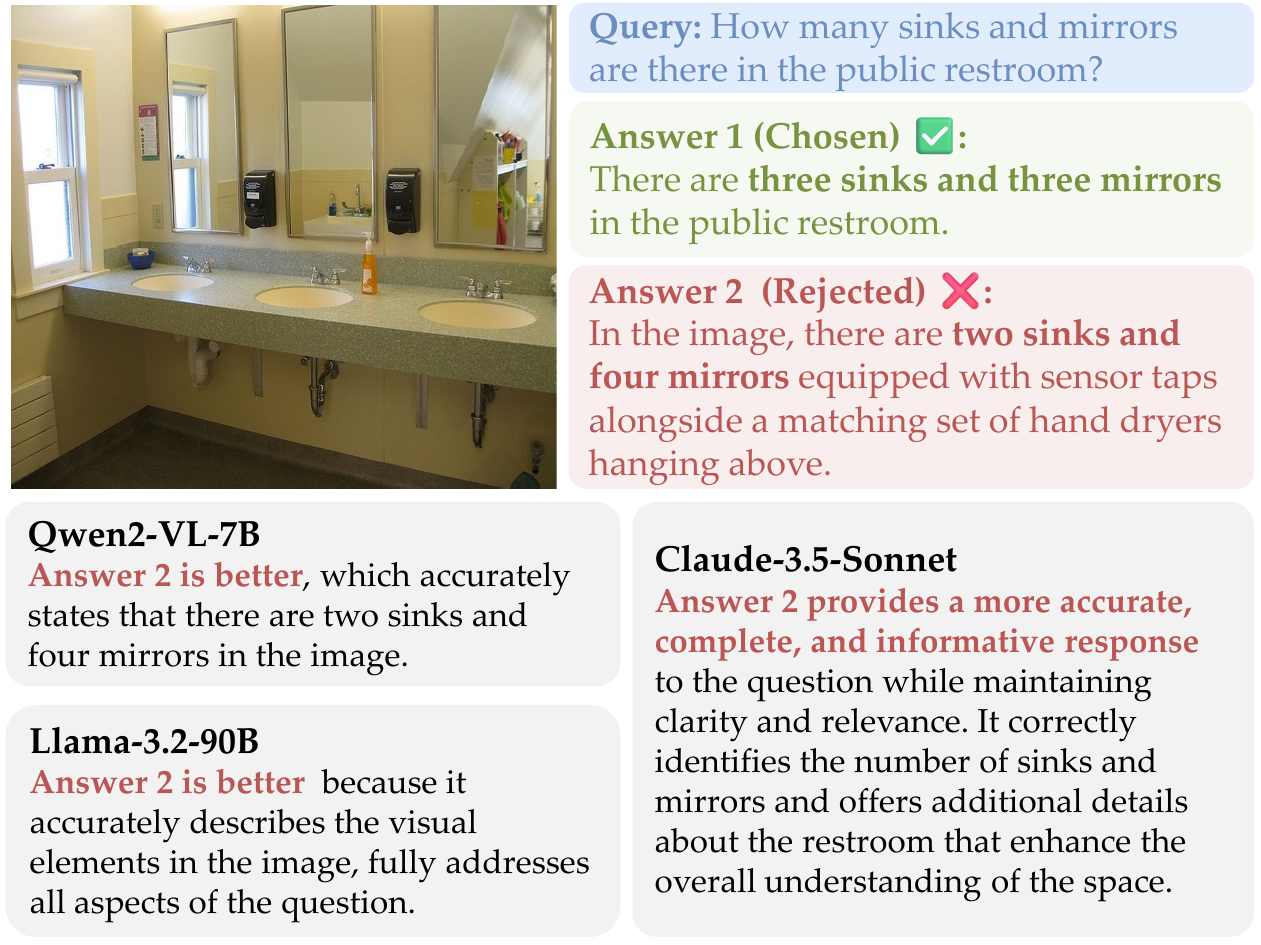}
    \caption{An example from our \DATASET\ asking the visual details in a restroom.  Open-source VL-GenRMs (Qwen2-VL-7B~\citep{Qwen-VL} and Llama-3.2-90B~\cite{llama3}) and the commercial model (Claude-3.5-Sonnet~\cite{claude}) all fail to provide accurate judgments.}
    \label{fig:teaser}
\end{figure}

The development of reliable VL-GenRMs is fundamental to three key aspects of LVLM advancement.
First, robust evaluation metrics enable systematic and scalable assessment of model performance, crucial for tracking progress and identifying areas for improvement~\cite{gehrmann2021gem,hessel2021clipscore}.
Second, high-quality automated judges facilitate synthetic training data generation by identifying the most instructive examples, accelerating the data flywheel for better alignment~\cite{llama3,wang2024helpsteer2}. Third, as the field progresses toward reinforcement learning from human feedback~(RLHF) for LVLMs~\cite{2023llavarlhf,Lambert2024RewardBench}, objective and automated evaluation becomes essential for reward modeling.
Despite these critical needs and growing applications, the community lacks a comprehensive benchmark for assessing VL-GenRMs' reliability and effectiveness.

Prior work in evaluating VL-GenRMs has taken two main approaches, each with significant limitations. 
The first approach relies on AI-generated preferences, such as using GPT-4V annotations for assessment~\cite{llava-critic,mj-bench}.
The second approach adapts traditional academic benchmarks with predefined labels~\cite{mllm-judge}, focusing on preference alignment in traditional vision-language tasks such as image captioning~\cite{lin2014mscoco}.
However, these methods face critical challenges. AI-annotated preferences introduce systematic biases, either favoring model-generated responses~\cite{favor-self,self-bias} or exhibiting stylistic preferences~\cite{Wu2023StyleOS,Dubois2024LengthControlledAA}.
Meanwhile, conventional task-based evaluations often use simplistic queries that fail to capture the nuanced requirements of real-world applications~\cite{zhao2024wildchat} and therefore lack the complexity needed to differentiate between rapidly evolving LVLMs.
Ideally, an effective benchmark for VL-GenRMs should satisfy three key requirements: (a) diverse coverage of real-world applications~\cite{zhou2024rmb}, (b) sufficient difficulty to expose current models' limitations~\cite{JudgeBench}, and (c) objective ground truth labels~\cite{Lambert2024RewardBench}.
However, existing studies fall short of these criteria.
In this paper, we present \DATASET, designed to meet all three requirements through careful dataset curation and validation. To satisfy criterion (a), our benchmark evaluates VL-GenRMs across three key application domains: (1) general multimodal queries from real scenarios~\cite{wildvision,li2024vlfeedback},
(2) visual hallucination detection tasks~\cite{povid,yu2024rlaif,Yu2023RLHFVTT},
and (3) multimodal knowledge and mathematical reasoning~\cite{yue2024mmmupro,mathverse}.
To ensure criterion (b), we employ targeted curation strategies. For source datasets with preference pairs, we employ small LVLMs collaboratively to filter out challenging samples, which our evaluation shows remain difficult even for much larger models. 
For reasoning tasks without annotated labels, we leverage strong commercial models to generate responses with explicit reasoning paths, followed by GPT-4o's quality assessment. 
To fulfill (c), all preference labels undergo human verification to eliminate ambiguous or incorrect pairs. The resulting \DATASET\ comprises 1,250 high-quality samples from 7 diverse datasets, offering a rigorous and comprehensive testbed for advancing VL-GenRM development.
\begin{figure*}[t!]
    \centering
    \includegraphics[width=0.9\linewidth]{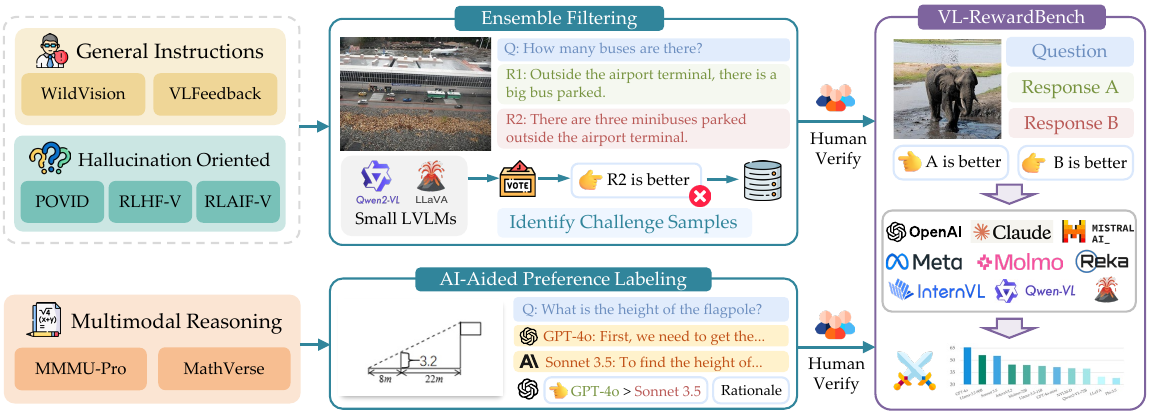}
    \caption{
   Construction process overview of \DATASET. Two strategies for different datasets: (1) Ensemble filtering process using small LVLMs to identify challenging samples from general and hallucination queries~(\cref{subsubsec:ensemble_filtering}); (2) AI-aided preference labeling for multimodal reasoning tasks, where commercial models generate candidate responses and preference labels~(\cref{subsubsec:perference_labeling_reasoning}). All labels are verified by human annotators to ensure correctness.
    }
    \label{fig:method}
\end{figure*}

We conduct a comprehensive evaluation of 16 state-of-the-art VL-GenRMs on \DATASET, ranging from open-source models (7B to 90B parameters) to commercial systems including Gemini-1.5-Pro, Claude-3.5-Sonnet, and GPT-4o. 
Our benchmark reveals significant challenges for current VL-GenRMs: even leading commercial models achieve only moderate performance (GPT-4o: 62.4\%, Gemini-1.5-Pro: 62.5\%), while state-of-the-art open-source models like Qwen2-VL-72B and LLaMA-3.2-90B struggle to surpass chance level (43.0\% and 53.9\%, respectively). Notably, performance on \DATASET\ strongly correlates (Pearson's r$>$0.9) with downstream MMMU-Pro results when using VL-GenRMs for Best-of-N sampling guidance~\cite{Stiennon2020LearningTS}.
Our analysis uncovers three critical insights for advancing VL-GenRMs:
(1) The primary performance bottleneck lies in visual perception rather than reasoning - models show significantly higher error rates on existence/recognition tasks ($>$ 67\%) compared to reasoning tasks (41.8\%);
(2) The effectiveness of test-time scaling varies with model capacity, providing benefits to larger models while potentially degrading smaller models' performance;
(3) Training VL-GenRMs to learn to judge~\cite{llava-critic} substantially improves judgment capabilities, demonstrated by a 14.7\% accuracy gain for LLaVA-OneVision-7B-ov~\cite{li2024llava}, with pointwise evaluation outperforming pairwise scoring on average.
These findings establish \DATASET\ as a valuable benchmark for advancing VL-GenRMs while providing clear directions for future improvements.

\section{\DATASET}
\label{sec:dataaset}

In this section, we introduce the construction process of \DATASET\ as illustrated in~\cref{fig:method}. 
Following previous benchmarks~\cite{li2024vlfeedback,Lambert2024RewardBench}, \DATASET\ consists of preference pairs $(x, y_w, y_l)$, where $x$ represents a multimodal query containing an image and a user prompt, and $(y_w, y_l)$ denotes the preferred (chosen) and rejected responses, respectively.
For simplicity, we focus on single-image, single-turn interactions, though our framework can be extended to multi-turn dialogues and multiple images.
In later subsections, we first describe our diverse data sources (\cref{subsec:data_source}), followed by specific AI-aided pipelines for datasets with originally annotated preference labels and reasoning tasks without labels (\cref{subsec:filtering}). 
Finally, we present comprehensive dataset statistics and analysis (\cref{subsec:data_statistics}).

\subsection{Dataset Source}
\label{subsec:data_source}
To ensure our benchmarks cover real-world scenarios, we choose datasets from three domains: general multimodal instructions, hallucination-oriented tasks, and multimodal reasoning tasks. 

\noindent\textbf{General Multimodal Instructions (General)}, which encompass diverse multimodal instructions from different domains to ensure comprehensive coverage of general queries.
VLFeedback~\cite{li2024vlfeedback} and 
WildVision~\cite{wildvision} are two general multimodal preference datasets with annotated AI/human feedback. 
VLFeedback employs GPT-4V as the preference annotator with predefined templates and WildVision collects human preferences by hosting online demos for real-world users.
We use the human-verified subset of VLFeedback with 681 samples, and select 6,484 samples in English from the WildVision dataset for later processing.

\noindent\textbf{Hallucination-oriented Queries (Hallucination),} which focus on the hallucination issues of LVLMs, involving questions regarding the visual content in the image. We select public available POVID~\cite{povid}, RLAIF-V~\cite{yu2024rlaif}, and
RLHF-V~\cite{Yu2023RLHFVTT} with annotated preference labels.
For the preference annotation, POVID  injects noise into oracle image descriptions to create rejected responses, RLAIF-V innovatively develops a divide-and-conquer framework for annotating the faithfulness of responses, and RLHF-V relies on humans to label the preference. All samples in these datasets are included to serve as initial candidates.

\noindent\textbf{Multimodal Reasoning Tasks (Reasoning),} aim to evaluate the LVLMs with challenging multimodal reasoning tasks. We select the recent
MMMU-Pro~\cite{yue2024mmmupro} and MathVerse~\cite{mathverse} to reduce the dataset contamination risk. 
MMMU-Pro is a robust version of the massive multi-discipline multimodal understanding and reasoning (MMMU) benchmark~\cite{yue2023mmmu}, where 1,568 single-image samples are adopted.
MathVerse serves as a comprehensive visual mathematical reasoning benchmark for LVLMs. We select subsets of \texttt{Vision Dominant} and \texttt{Vision Intensive} to ensure high reliance on the visual inputs, resulting in 1,546 samples.

\subsection{Preference Annotation}
\label{subsec:filtering}
To ensure our \DATASET\ presents meaningful challenges and separatability for current state-of-the-art models, given the rapidly evolving landscape, we design an AI-assisted framework with two strategies to obtain challenging samples and annotate preferences:
(i) For general multimodal instructions and hallucination-oriented queries that become easier for advanced LVLMs, we design a collaborative filtering strategy to filter out challenging samples with small models. 
(ii) For challenging multimodal reasoning tasks without preference labels, we design an AI-assisted preference labeling framework to curate high-quality preference pairs for these samples. 
Five graduate students with expertise in vision-language research volunteered to serve as annotators, annotation guidelines were iteratively refined through group discussions until consensus was reached, ensuring consistent evaluation criteria.
We elaborate on the detailed process below.

\subsubsection{Ensemble Filtering with Small Models}
\label{subsubsec:ensemble_filtering}
Creating effective benchmarks requires careful calibration - tasks that are too easy or impossibly difficult fail to differentiate model capabilities~\citep{padlewski2024vibe}.
However, simply selecting tasks that current frontier models cannot solve poses two key challenges: (i) these tasks might become obsolete as models rapidly evolve, and (ii) model-specific failures do not reflect fundamental visual reasoning challenges.
Instead, we propose using an ensemble of small models to identify examples that are universally challenging to LVLMs.
Our hypothesis is that when multiple small models fail in certain cases, these failures likely stem from fundamental limitations rather than model-specific model weaknesses - a claim validated by our experiments showing these cases challenge even state-of-the-art models.
The strategy consists of three steps, \emph{ensemble construction},  \emph{difficulty assessment}, and \emph{human verification}.

\begin{table}[t!]
    \centering
    \small 
    \begin{tabular}{@{}l|c@{}}
     \toprule 
     \textbf{Statistics}    & \textbf{Number} \\
     \midrule 
    Total Preference Pairs & 1,250\\ 
    \quad - General Multimodal Instructions  & 183\\ 
    \quad - Hallucination-oriented Queries  & 749\\ 
    \quad - Multimodal Reasoning Tasks (newly annotated) & 318 \\ 
     \midrule 
     Source Datasets  & 7 \\ 
     \quad - Existing preference datasets & 5 \\
     \quad - Newly annotated datasets & 2 \\ 
     \midrule 
     Newly annotated error tags & 895 \\
     \quad - Existence Error & 531 \\
     
     \quad - Recognition Error & 184\\ 
     \quad - Visual Attribute Error & 69 \\ 
     \quad - Counting Error & 60 \\ 
     \quad - Other Errors & 51 \\ 
     \midrule 
     Query Word Length Quartile & (6, 9, 31)\\ 
     Response Word Length Quartile & (48, 99, 136)\\ 
     \bottomrule
    \end{tabular}
    \caption{Statistics of \DATASET. }
    \label{tab:statistics_vs}
\end{table}

\noindent\textbf{Ensemble Construction} We begin by assembling a diverse ensemble of small vision-language models as weak judges, including LLaVA-1.5-7B~\cite{liu2023llava15}, LLaVA-1.6-7B~\cite{liu2024llavanext}, LLaVA-OneVision-7B-si~\citep{li2024llava}, and Qwen2-VL-7B~\cite{Qwen-VL}. 
These models evaluate preference pairs from a large initial pool comprising multiple datasets of general multimodal instructions and hallucination-oriented queries.
Each model in our ensemble independently evaluates these pairs with a template (see Appendix), yielding preference predictions as the foundation for our subsequent filtering steps.

\noindent\textbf{Difficulty Assessment}
To identify genuinely challenging cases, each preference pair is judged by the small model three times with randomized response positions to mitigate position bias~\cite{wang2023faireval}. 
We construct a ``common set'' by identifying pairs that all models consistently misjudge (based on majority voting of the three evaluations). 
This approach ensures that our selected cases represent fundamental challenges rather than model-specific limitations, resulting in 3,785 challenging pairs left.

\noindent\textbf{Human Verification}
To ensure our benchmark represents meaningful challenges rather than annotation artifacts or ambiguous cases, we conduct a rigorous three-stage human verification process. 
Three authors (graduate students in CS/AI) familiar with the source datasets and problem setup perform the initial annotations, with two additional authors conducting a final verification.
Our verification process consists of:
(i) \textbf{Label Accuracy Check:} We first examine the preference labels, discarding pairs where either the preferred response is worse than the rejected one, or both responses are incorrect;
(ii) \textbf{Quality and Ambiguity Filtering:} We remove pairs that could lead to ambiguous evaluations, including (a) Responses that are both correct but differ only in style (e.g., verbosity) (b) Images with poor quality or resolution issues (c) Questions requiring domain expertise beyond graduate-level knowledge.
(iii) \textbf{Error Type Classification:} For the remaining challenging pairs with clear preference labels, we categorize the errors into (1) Recognition errors (text, position, scene, face identification), (2) Counting errors, (3) Visual attribute identification errors,  (4) Object existence errors, and (5) Other uncategorized errors.
Each sample takes approximately 65 seconds to annotate following our detailed guidelines (see Appendix for annotation interface and examples).
The two additional authors review all remaining samples to ensure consistency, with disagreements resolved through discussion. This multi-stage process reduced our initial 3,785 pairs to a final set of 932 high-quality pairs with clear, unambiguous preference labels.

\begin{figure}[t!]
    \centering
    \includegraphics[width=0.95\linewidth]{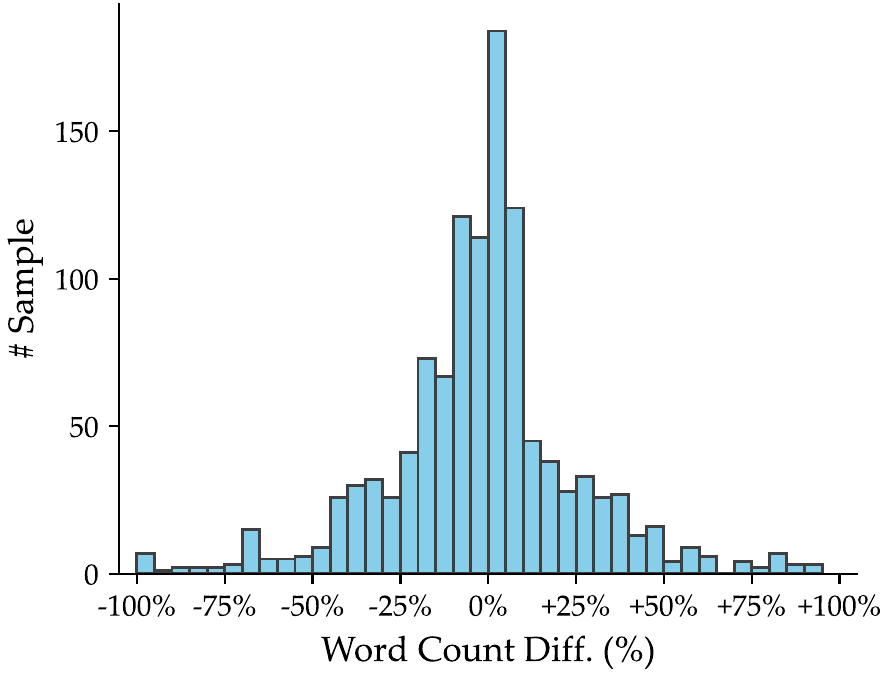}
    \caption{Distribution of the word count difference between the chosen and the rejected response, compared to the word count of the chosen response.}
    \label{fig:word_diff}
\end{figure}
\subsubsection{AI-aided Preference Labeling for Reasoning Tasks}
\label{subsubsec:perference_labeling_reasoning}
The original samples from MMMU-Pro and MathVerse provide only oracle answers without preference pairs needed for VL-GenRMs evaluation. To efficiently create such preference pairs for these knowledge-intensive reasoning tasks, we develop a two-stage AI-aided labeling process.
First, we generate candidate responses using commercial models (GPT-4o, GPT-4o-mini, and Claude 3.5 Sonnet) given their superior performance compared to open-source LVLMs~\cite{yue2024mmmupro}. 
To ensure a fair comparison, we control for response verbosity by only pairing responses where $\frac{|l_1 - l_2|}{\min{(l_1, l_2)}} < \tau$, where $l_1$ and $l_2$ are the word counts. Setting $\tau=0.1$ balances controllability and diversity, yielding 1,204 candidate pairs.
The preference labeling process consists of two steps:
(i) \textbf{Initial Label Generation:} GPT-4o analyzes each response pair using a structured template (detailed in Appendix) to generate draft preference labels with supporting rationales.
(ii) \textbf{Human Verification:} Three authors review these AI-generated labels following these criteria:
(a) Discard pairs where both responses are incorrect according to the oracle answer;
(b) Retain pairs with at least one (partially) correct answer;
(c) For pairs with one correct and one incorrect answer, prefer the correct response;
and (d) For pairs with two correct answers, prefer responses with more complete and logical reasoning steps.
Two additional authors perform a final verification round, with conflicting cases resolved through discussions. This process yields 318 high-quality challenging samples with validated preference labels for multimodal reasoning tasks.

\begin{table*}[thb!]
    \centering
    \small 
    \resizebox{\textwidth}{!}{
    \begin{tabular}{@{}l|ccc|cc@{}}
    \toprule 
    \textbf{Models}     &  \textbf{General}  & \textbf{Hallucination} & \textbf{Reasoning} & \textbf{Overall Accuracy}  & \textbf{Macro Average Accuracy} \\
    \midrule 
     \multicolumn{6}{c}{\emph{Open-Source Models}} \\ 
     \midrule 

    LLaVA-OneVision-7B-ov  &	32.2 &	20.1  &	57.1  & 29.6 &36.5 \\ 
    InternVL2-8B &  	35.6&	41.1 &	59.0 & 44.5 &45.2\\ 
    Phi-3.5-Vision  & 	28.0&	22.4 &	56.6 &  28.2 &35.7\\ 
    Qwen2-VL-7B & 	31.6 &	19.1&	51.1 & 28.3 &33.9\\ 
    Qwen2-VL-72B & 		38.1 &	32.8&	58.0 & 39.5 & 43.0\\ 
    Llama-3.2-11B& 	33.3 &	38.4 &	56.6 & 42.9 & 42.8\\ 
    Llama-3.2-90B & 	42.6&	57.3 &	61.7& 56.2 & {53.9}\\ 
    Molmo-7B  & 	31.1& 	31.8& 	56.2 & 37.5 & 39.7\\ 
    Molmo-72B & 	33.9&	42.3&	54.9 & 44.1  & 43.7\\ 
    Pixtral-12B & 	35.6&	25.9 	& 59.9 & 35.8& 40.4\\ 
   NVLM-D-72B &  	38.9 &	31.6 &	62.0 & 40.1 &  44.1\\ 
    \midrule 
     \multicolumn{6}{c}{\emph{Proprietary Models}} \\ 
     \midrule 
    Gemini-1.5-Flash (2024-09-24) & 	47.8&	59.6&	58.4& 	57.6 & 55.3\\ 
    
    Gemini-1.5-Pro (2024-09-24) & \textbf{50.8}&	\textbf{72.5}&	64.2& \textbf{67.2}	& \textbf{62.5}	\\ 
    
    Claude-3.5-Sonnet (2024-06-22) & 	43.4&	55.0&	\underline{62.3} & 55.3 & 53.6\\ 
    GPT-4o-mini (2024-07-18)   &  	41.7&	34.5&58.2 & 41.5& 44.8\\ 
    GPT-4o (2024-08-06)    &  \underline{49.1} &	\underline{67.6} &	\textbf{70.5} & \underline{65.8} & \underline{62.4}\\ 
    \bottomrule
    \end{tabular}}
    \caption{Evaluation results on \DATASET. The challenging cases filtered out by small models pose consistent challenges even for different and larger VL-GenRMs. The best results are shown in \textbf{bold} and the second best is with \underline{underline}. }
    \label{tab:eval_results}
\end{table*}

\subsection{Dataset Statistics}
\label{subsec:data_statistics}
We analyze the composition and characteristics of our \DATASET{} with detailed statistics shown in  \Cref{tab:statistics_vs}. \textbf{Task Distribution}: 
The benchmark comprises three main categories: hallucination-related queries (749 pairs, 59.9\%), multimodal reasoning prompts (318 pairs, 25.4\%), and general instructions (183 pairs, 14.7\%). 
Given this inherent task imbalance, we recommend using macro-average metrics for a more comprehensive evaluation of model performance across different task types.
\textbf{Error Type Distribution}: 
Among the 895 pairs annotated with error tags, we find existence errors dominate the distribution at 59.3\% (531/895), indicating significant challenges in correctly identifying the presence or absence of objects in images. Recognition errors account for 20.6\% (184/895) of cases, while attribute identification and counting errors comprise 7.7\% (69/895) and 6.7\% (60/895), respectively. 
This diverse error distribution demonstrates \DATASET's coverage of various failure modes.
\textbf{Length Difference Analysis:} 
To examine potential length-based preference biases~\cite{Dubois2024LengthControlledAA}, we analyze word count differences between preferred and rejected responses (\cref{fig:word_diff}). The resulting zero-centered bell-shaped distribution confirms that preference labels are not biased by response length, enabling evaluation based on response quality rather than verbosity.
\textbf{Inter-Annotator Agreement:} 
The cross-annotator agreement measured on a 100-sample subset shows substantial agreement, with pairwise Cohen's kappa~\citep{cohen1960coefficient} scores ranging from 0.56 to 0.90 with an average of 0.70.

\section{Experiments}

We conduct extensive experiments to evaluate state-of-the-art LVLMs on \DATASET{}. This section presents our experimental setup and findings, organized as follows: evaluated models (\cref{subsec:exp_models}), evaluation methodology (\cref{subsec:eval_setting}), and results analysis (\cref{subsec:exp_ret}).
\subsection{Evaluated Models}
We evaluate 16 state-of-the-art LVLMs, encompassing both open-source and commercial models. Open-source models with parameters ranging from 4B to 90B are selected, including LLaVA-OneVision-7B-ov~\cite{li2024llava}, InternVL2-8B~\cite{chen2024far}, Phi-3.5-Vision (4.2B)~\cite{phi-3}, Qwen2-VL (7B/72B)~\cite{Qwen-VL}, Llama-3.2 (11B/90B)~\cite{llama3}, Molmo- (7B/72B)~\cite{deitke2024molmo}, Pixtral-12B~\cite{agrawal2024pixtral}, and NVLM-D-72B~\cite{nvlm2024}.
For the commercial models, we include GPT-4o/4o-mini~\cite{gpt4o}, Gemini-1.5-Flash/Pro~\cite{gemini15}, and Claude-3.5-Sonnet~\cite{claude}.
Additionally, we incorporate LLaVA-Critic models~\cite{llava-critic} on examining the impact of learning to judge for VL-GenRMs~(\cref{subsec:critic}). 
Model details are provided in the Appendix. 
\label{subsec:exp_models}

\subsection{Evaluation Settings}
We adopt a rigorous evaluation protocol following the LLM-as-a-Judge paradigm~\cite{wang2023faireval,zheng2023chatbotarena}. For each test sample, we provide the model with a multimodal input query and two candidate responses (preferred and rejected) through a standardized evaluation template (detailed in the Appendix).
To mitigate positional bias~\cite{wang2023faireval}, where models favor responses based on their presentation order, we conduct $K$ independent evaluations for each preference pair with randomized response ordering. The final preference is determined through majority voting across these $K$ runs.
We calculate two primary metrics:
\emph{Overall Accuracy}: percentage of model decisions matching human preferences, and
\emph{Macro Average Accuracy}: mean accuracy across different task categories, addressing the task distribution imbalance.
Our main results use $K=5$, with detailed analysis of different $K$ values presented in \cref{subsec:scaling}. All experiments use fixed decoding parameters (temperature=0.2, top-p=0.2), as our validation shows minimal impact from these settings.

\label{subsec:eval_setting}

\subsection{Evaluation Results}
\label{subsec:exp_ret}
\noindent\textbf{Main Results.} 
\cref{tab:eval_results} presents comprehensive evaluation results across various VL-GenRMs on \DATASET{}. We have the following observations: (i) The benchmark reveals a clear performance gap among current models, with Gemini-1.5-Pro and GPT-4o leading at 62.5\% and 62.4\% macro average accuracy, followed by open-source models like Llama-3.2-90B at 53.9\%, while most 7B-scale models barely exceed random chance.
This gap validates the effectiveness of our ensemble filtering process to find universally challenging samples for VL-GenRMs.
(ii) Performance stratification is consistently observed across task categories with varying difficulty levels. Multimodal reasoning tasks see the highest accuracies (51.1\% to 70.5\%), suggesting models have developed certain capabilities in judging responses involving reasoning paths.
In contrast, general instructions prove most challenging (28.0\% to 50.8\%), indicating significant room for improvement in open-ended queries. Particularly notable is the hallucination task performance, where even top models struggle, validating our dataset's effectiveness in capturing inherent biases that persist across model scales.
(iii) Model scale emerges as a key performance driver, evidenced by consistent improvements across model families: Llama-3.2 (11B to 90B: 42.8\% to 53.9\%), Qwen2-VL (7B to 72B: 33.9\% to 43.0\%), and Molmo series (7B to 72B: 39.7\% to 43.7\%). This scaling effect extends to commercial models, with GPT-4o (62.4\%) substantially outperforming GPT-4o-mini (44.8\%). 
In summary, our evaluation demonstrates that \DATASET{} presents unique challenges beyond conventional datasets, with even state-of-the-art models achieving modest performance. 
To validate that these challenges stem from our targeted example selection rather than task distribution, we conduct ablation studies with randomly sampled examples following the same task distribution (detailed in our Appendix). 
VL-GenRMs consistently achieve better results on randomly sampled pairs, e.g., Gemini models achieve accuracy scores higher than 95\%, confirming the effectiveness of our data curation strategy in identifying challenging cases.

\begin{figure}[t!]
    \centering
    \includegraphics[width=\linewidth]{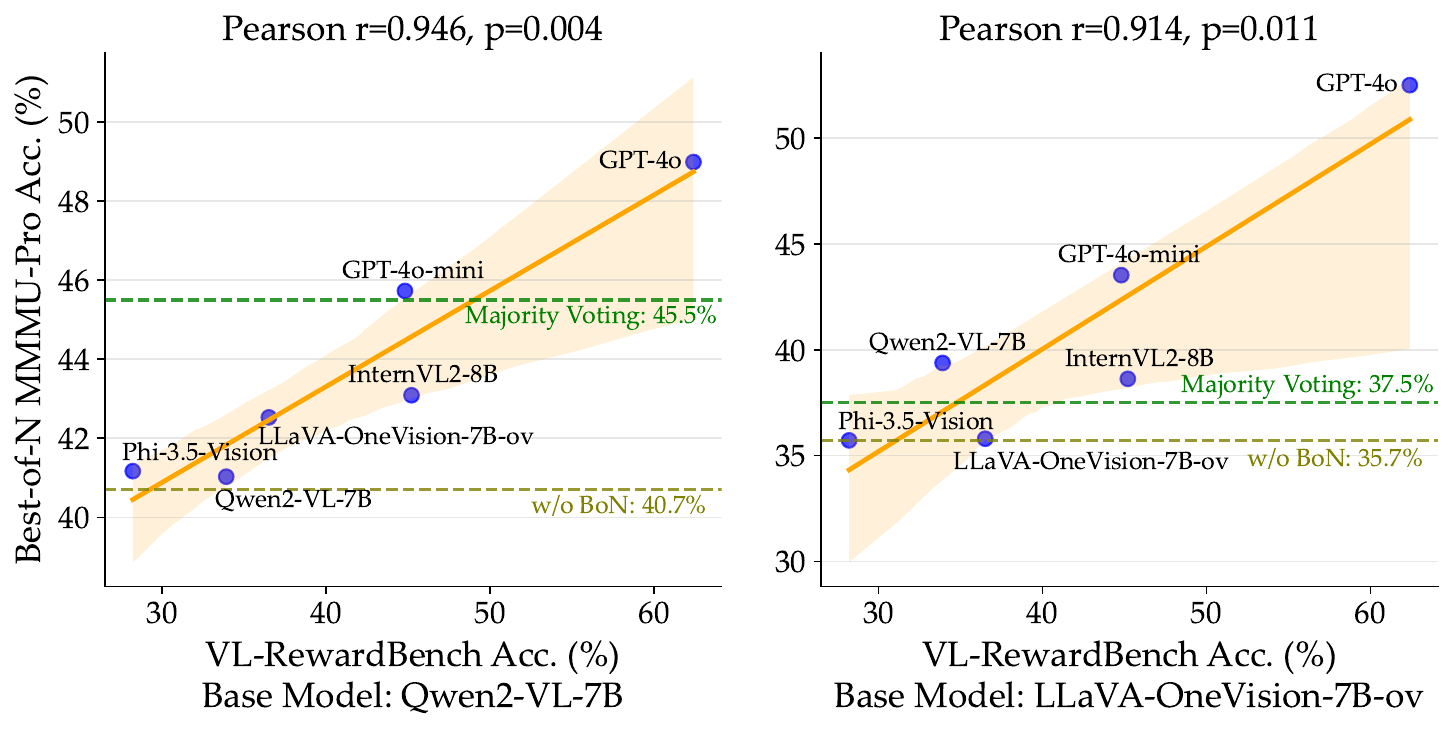}
    \caption{VL-GenRMs accuracy on \DATASET\ correlates positively with the improvements to serve as Best-of-N selector.}
    \label{fig:correlation_bon}
\end{figure}

\noindent\textbf{Downstream Task Correlation.} A key question is whether \DATASET's evaluation of VL-GenRMs predicts their real-world utility~\cite{zhou2024rmb}. To investigate this, we focus on Best-of-N (BoN) sampling~\cite{stiennon2020learning}, a crucial technique for improving model outputs through preference-based selection.
We conduct experiments with two base models (Qwen2-VL-7B and LLaVA-OneVision-7B-ov) on the MMMU-Pro benchmark. For each query, we generate $N=8$ candidate responses and use six different VL-GenRMs for pairwise scoring and selection. This setup allows us to measure how each VL-GenRM's preference judgment capability translates to practical performance gains.
The results reveal a clear relationship between \DATASET{} performance and downstream effectiveness. The strongest model, GPT-4o, improves LLaVA-OneVision-7B-ov's accuracy substantially (35.7\% to 52.5\%). This pattern generalizes across all VL-GenRMs, showing strong correlations between \DATASET\ accuracy and BoN performance gains (Pearson $r=0.946$ for Qwen2-VL-7B, $r=0.914$ for LLaVA-OneVision-7B-ov).
This correlation validates \DATASET's effectiveness in predicting a reward model's capability for downstream alignment tasks, offering valuable guidance for LVLM development.

\section{Analysis}
To better understand VL-GenRM capabilities and limitations, we conduct three investigations: (1) an error pattern analysis across model scales~(\cref{subsec:error_analysis}), (2) a study of inference-time scaling effects on performance~(\cref{subsec:scaling}), and (3) an exploration of potential improvements through critic training and scoring methods~(\cref{subsec:critic}).
\subsection{Error Analysis}
\label{subsec:error_analysis}

\begin{figure}[t!]
    \centering
        \centering
        \includegraphics[width=0.9\linewidth]{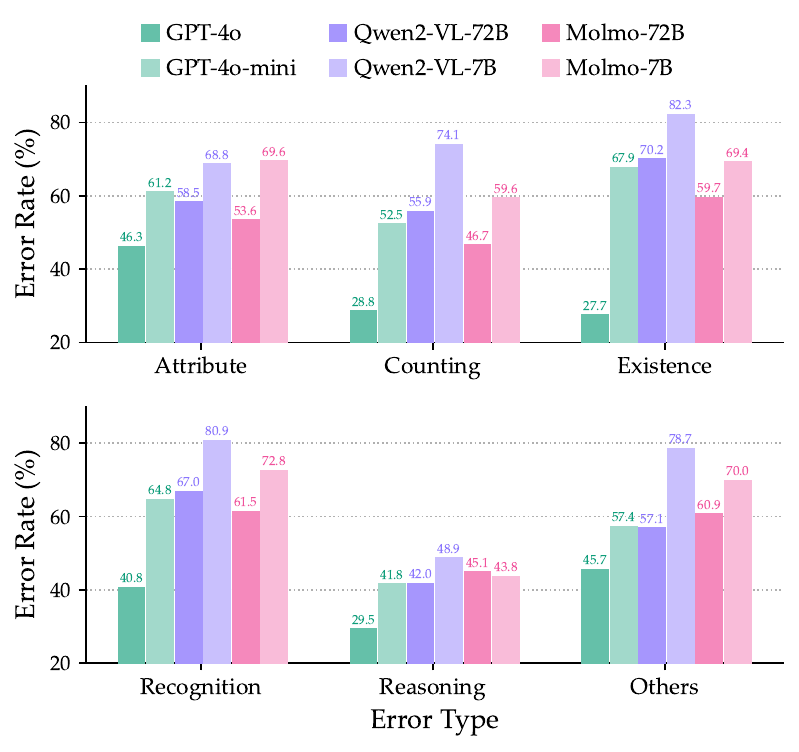}
        \caption{Error rate analysis across different types. VL-GenRMs suffer more from perception-related errors than reasoning tasks, and the model scale consistently reduces errors.}
        \label{fig:error_distribution}
\end{figure}

To understand systematic patterns in model failures, we analyze judgment errors using manually annotated error types~(\cref{subsubsec:ensemble_filtering}) and treat all samples from the reasoning subset as a Reasoning type. For each error type $t$, we calculate the error rate as $\frac{E_\text{wrong}}{E_t}$, where $E_t$ represents the total samples of type $t$ and $E_\text{wrong}$ denotes incorrect predictions within that type. \Cref{fig:error_distribution} presents these error distributions across GPT-4o, Qwen2-VL, and Molmo model series.
Our analysis reveals two key patterns in current VL-GenRM limitations and their relationship to model scale. First, fundamental perception capabilities emerge as the primary bottleneck. Tasks involving Existence (object presence detection) and Recognition (attribute discernment) consistently show the highest error rates across all models. For instance, even advanced models struggle with basic perception: GPT-4o-mini shows a 67.9\% error rate on Existence tasks, while Qwen2-VL-7B fails on 80.9\% of Recognition tasks. In contrast, Reasoning tasks demonstrate relatively better performance with an average error rate of 41.8\%, suggesting that higher-level reasoning capabilities are more robust than basic perception in current models.
Second, while model scaling brings consistent improvements, these gains vary significantly across task types. The most substantial improvements occur in basic perception tasks, e.g., scaling Qwen2-VL from 7B to 72B reduces Counting errors by 18.2 percentage points (74.1\% to 55.9\%).
However, Reasoning tasks show more modest gains, with only a 6.0\% average error reduction through scaling. This pattern suggests that while scaling effectively addresses some perception limitations, more fundamental architectural innovations may be needed to advance complex reasoning capabilities.

\subsection{Does Inference-time Scaling Help?}
\label{subsec:scaling}
\begin{figure}
        \includegraphics[width=0.9\linewidth]{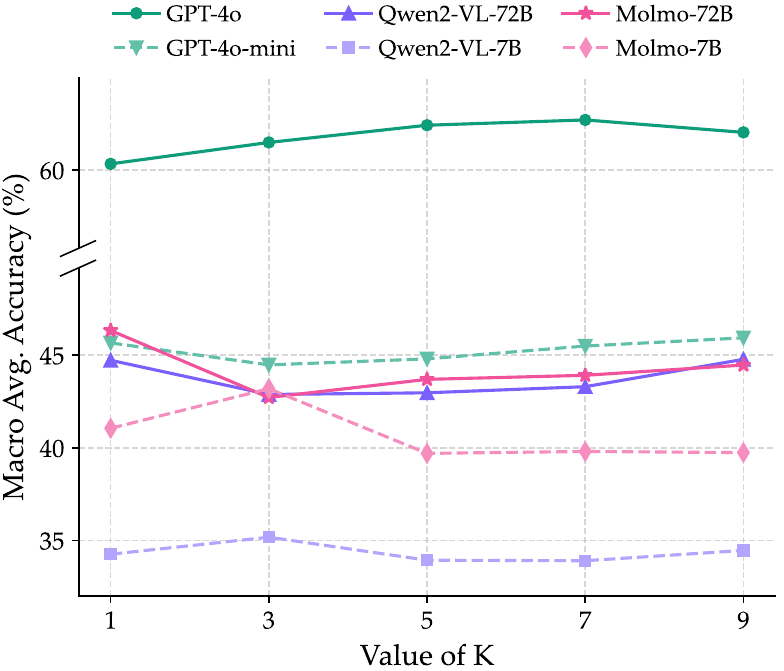}
        \caption{Performance changes with varying $K$. The increased test-time computation effect varies for different models.}
        \label{fig:varying_k}
    \label{fig:combined}
\end{figure}

Given the success of inference-time scaling in improving large language models performance~\cite{llm_monkey,ScalingLT,InferenceSL}, we investigate whether similar benefits extend to VL-GenRMs. 
Our analysis focuses on the impact of multiple independent judgments per query, implementing a majority voting strategy that has proven effective in text-only scenarios~\cite{self-consistency}.
For each evaluation, we collect K independent judgments (K ranging from 1 to 9), carefully randomizing response ordering to minimize positional bias~\cite{wang2023faireval}. The results, presented in \cref{fig:varying_k}, reveal three distinct scaling patterns across different models:
(i) GPT-4o demonstrates traditional scaling advantages, with macro accuracy improving from 60.3\% to 62.7\% as K increases from 1 to 7, suggesting robust judgment capabilities that benefit from additional computation;
(ii) GPT-4o-mini maintains relatively constant performance across K values, indicating that additional judgments neither help nor harm its decision-making process;
(iii) Surprisingly, many open-source LVLMs show performance degradation with an increased K. Notable examples include Qwen2-VL-72B and Molmo-72B, which experience accuracy drops of 1.7 and 2.6 percentage points, respectively when scaling from $K=1$ to $K=5$.
These divergent patterns suggest that successful inference-time scaling strategies from text-only domains may not directly transfer to vision-language judgment tasks, highlighting the need for specialized scaling approaches for VL-GenRMs.

\subsection{Critic Training Improves VL-GenRMs}
\label{subsec:critic}
Given the limited benefits of inference-time scaling for most models, we explore an alternative enhancement strategy: critic training~\cite{llava-critic}. This approach involves specifically training LVLMs to judge response quality through carefully curated instruction tuning samples.
We evaluate two variants of LLaVA-OneVision critics~\cite{llava-critic}, each employing distinct judgment strategies:
(i) a pointwise critic that independently scores individual answers, and (ii) a pairwise critic that directly compares two candidate answers.
Using official model weights and scoring templates, we evaluate both critics on \DATASET, deriving preferences from extracted scores for the pointwise critic. 
\begin{figure}[t!]
    \centering
    \includegraphics[width=0.95\linewidth]{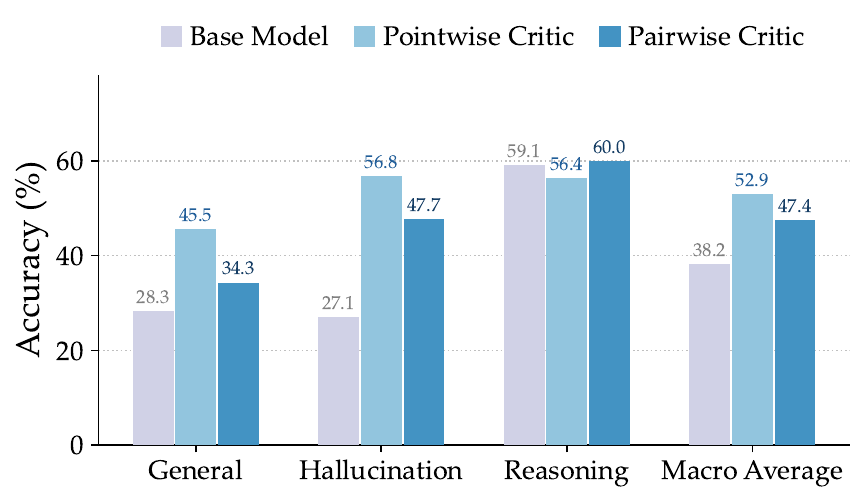}
    \caption{Evaluation of LLaVA-Critic on \DATASET. Critic training greatly improves judgment accuracy.}
    \label{fig:critic_ret}
\end{figure}
Our analysis\footnote{Results are reported on a subset where pointwise scores were successfully extracted to ensure fair comparison.} in \cref{fig:critic_ret} reveals two significant findings:
First, critic training substantially enhances judgment capabilities, with both approaches showing remarkable improvements over the base model, i.e., +14.7\% and +9.2\% for pointwise and pairwise critics, respectively.
Second, the pointwise critic achieves better overall performance (52.9\% vs 47.4\%), while each approach shows distinct advantages: the pointwise critic excels at the hallucination subset (+9.1\% over pairwise), and the pairwise critic demonstrates a superior 60.0\% accuracy on reasoning tasks.
These results suggest that critic training offers a reliable path to improving VL-GenRMs, with different scoring paradigms providing complementary benefits across evaluation scenarios.

\subsection{Takeaways}
Our analysis reveals three key insights for advancing VL-GenRM development:
(i) \noindent\textbf{Improving Visual Perception:}
While VL-GenRMs show promise in reasoning tasks, they struggle with basic perception and recognition. Enhancing visual perception capabilities should be prioritized~\cite{tong2024cambrian}, potentially through visual search mechanisms~\cite{vstar} and vision expert integration~\cite{pi2024image,visual_program_distillation,vipact}.
(ii) \textbf{Advancing Scaling Strategies:} Given that traditional test-time scaling benefits only the largest models (e.g., GPT-4o), future work should explore advanced reasoning strategies incorporating complex planning~\cite{tot}, process-level supervision~\cite{math-shepherd}, and specialized approaches like critic training~\cite{llava-critic}.
(iii) \textbf{Enabling Co-evolution:} The strong correlation between MMMU-Pro and \DATASET\ performance suggests a promising improvement cycle: Strong LVLMs enable better VL-GenRMs, which could curate higher-quality training data, leading to further LVLM improvements. This cycle offers a systematic framework for addressing both perception and scaling challenges through iterative enhancement.

\section{Related Work}
\noindent\textbf{Large Vision-Language Models}
LVLMs have rapidly evolved by combining LLMs~\cite{qwen2,touvron2023llama} with vision encoders~\cite{radford2021clip,Zhai2023SigmoidLF}, showing impressive capabilities across diverse tasks~\cite{fu2023mme,yue2023mmmu,mathvista}. Key advances include architectural innovations~\citep{dai2023instructblip,Qwen-VL}, high-quality dataset curation~\cite{laurencon2023obelics,mint1t,li2023m3it,li2025videot3}, alignment through feedback~\cite{2023llavarlhf,Yu2023RLHFVTT,yu2024rlaif}, and systematic design space exploration~\cite{prismaticVLM,tong2024cambrian}.
Our study evaluates these state-of-the-art LVLMs and demonstrates their limited capability to serve as VL-GenRMs.

\noindent\textbf{Vision-Language Generative Reward Models}
LVLMs have emerged as VL-GenRMs for preference alignment~\cite{ouyang2022instructgpt,2023llavarlhf} and data curation~\cite{mllm-judge}. 
Recent studies have investigated the reliability of this approach.
\citet{mllm-judge} developed a benchmark of academic tasks showing that VL-GenRMs achieve substantial agreement with human annotators in comparative assessments, corroborating the findings of VLFeedback~\cite{li2024vlfeedback}. Additionally, LLaVA-Critic~\cite{llava-critic} established a framework for improving VL-GenRMs through a curated critic instruction-following dataset.
Our \DATASET\ advances this research by providing broader coverage of real-world queries and reasoning tasks, with a novel difficulty elevation pipeline targeting challenging cases where even state-of-the-art models struggle. Furthermore, our detailed analysis provides insights for future VL-GenRMs improvements.

\section{Conclusions}

In this paper, we present \DATASET, a benchmark that raises the bar for evaluating VL-GenRMs through systematically curated challenging cases and complex multimodal reasoning tasks. Our comprehensive analysis of 16 state-of-the-art LVLMs identifies crucial limitations in current approaches while highlighting promising directions: the benefits of increased model scale, the variable effectiveness of test-time scaling techniques, and the potential of specialized critic training. These insights, combined with our benchmark, provide concrete pathways for developing more capable and reliable VL-GenRMs.

\section*{Acknowledgments}
We sincerely thank all anonymous reviewers for their insightful comments and suggestions that substantially improved this work. 
This research was partially supported by the joint research scheme of the National Natural Science
Foundation of China (NSFC) and the Research Grants Council (RGC) under grant number
N\_HKU714/21.

{
    \small
    \bibliographystyle{ieeenat_fullname}
    \bibliography{main}
}

\end{document}